\documentclass[runningheads]{llncs}

\usepackage{eccv}

\usepackage{eccvabbrv}

\usepackage{graphicx}
\usepackage{booktabs}
\usepackage{booktabs}
\usepackage{multirow}
\usepackage{caption} 
\usepackage[accsupp]{axessibility}  

\usepackage[pagebackref,breaklinks,colorlinks]{hyperref}
\usepackage{multirow}
\usepackage{xcolor}
\usepackage[table]{xcolor}
\newcommand{\blue}[1]{{\color{blue} }}
\usepackage{orcidlink}

\usepackage{tcolorbox}
\newtcolorbox{promptbox}[2][Prompt]{
  colback=black!5!white,
  arc=5pt,
  boxrule=0.5pt,
  fonttitle=\bfseries,
  title=#1,
  before upper={\small}, fontupper=\fontfamily{ptm}\selectfont,
  colframe=#2, 
}

\begin{document}

\title{Metric-Bench: Exploring In-context Spatial Metric Reasoning in VLMs for Indoor Scenes} 

\titlerunning{Metric-Bench}

\author{Yuling Xi\inst{1}$^\dag$\orcidlink{0009-0002-1420-7849} \and
Haokai Zhang\inst{1}$^\dag$ \and Muzhi Zhu\inst{1} \and Hao Zhong\inst{1} \and  Zongze Du\inst{1} \and  Hengyu Zhao\inst{1} \and  Chenchen Jing\inst{2} \and   Yufei Yin\inst{3}\orcidlink{0000-0001-9643-4601} \and  Bin Qin\inst{4} \and  Yongjie Yang\inst{4} \and  Zhenbo Luo\inst{4}\orcidlink{0009-0002-5836-0749} \and  Hao Chen\inst{1}\thanks{Corresponding author}\orcidlink{0000-0003-4417-614X} \and  Chunhua Shen\inst{1}\inst{2}\orcidlink{0000-0002-8648-8718} }
\authorrunning{Y.~Xi et al.}

\institute{State Key Lab of CAD \& CG, Zhejiang University, China \and Zhejiang University of Technology \and Hangzhou Dianzi University, China \and
Xiaomi Corporation, China\\
\url{https://huggingface.co/datasets/yulingxi/Metric-Bench}
}

\maketitle

\begin{abstract}
Metric reasoning is a critical and challenging task for Vision Language Models (VLMs), playing a pivotal role in embodied AI tasks such as robotic manipulation and autonomous navigation. However, current spatial reasoning remains bottlenecked by rigid pixel-level supervision; such localized optimization often compromises general multimodal intelligence, triggering performance degradation or catastrophic forgetting of broad reasoning capabilities. To address these limitations, we introduce Metric-Bench, a focused benchmark designed to guide metric-spatial reasoning using contextual information. By incorporating in-image reference objects with known physical dimensions, Metric-Bench guides models to implicitly learn the 2D-to-3D mapping without camera intrinsics. 
We further present MetricReasoner, a task-adapted reinforcement fine-tuning recipe for reference-grounded metric reasoning, using structured prompts and verifiable numerical rewards.
Extensive experiments on Metric-Bench demonstrate that our approach significantly enhances spatial metric understanding, outperforming existing and even larger proprietary models by 43.1\%, while improving downstream embodied performance over a spatial-specialized counterpart by 30.4\% on RoboSpatial overall accuracy and 9.3\% on ERQA, and additionally delivering consistent gains on general benchmarks (15.9\% on V$\star$Bench, 88.9\% on BLINK), indicating that the proposed adaptation does not necessarily compromise general VLM capabilities.
  \keywords{Vision language models \and Spatial reasoning  \and Metric measurement \and ScanNet}

\end{abstract}

\section{Introduction}
\label{sec:intro}

Understanding metric scale, the ability to infer real-world sizes and distances from images, is a fundamental aspect of visual intelligence, yet remains challenging for current Vision Language Models (VLMs).
While recent VLMs \cite{comanici2025gemini, Qwen2.5-VL,vteam2025glm45vglm41vthinkingversatilemultimodal,2025gpt5,Qwen3-VL} excel in semantic reasoning and instruction following, they often fail to make physically consistent judgments about spatial scale, such as estimating object size or distance \cite{yang2025thinking, daxberger2025mm, cai2025has, cai2025depthlm,liu2025ssr}. This gap reveals a missing bridge between semantic perception and metric reasoning, limiting their utility in downstream tasks such as navigation and manipulation.

Humans, by contrast, can estimate metric quantities by referencing known objects and logically inferring spatial relationships \cite{chen2022big,konkle2012familiar,gogel1987familiar,maruya2020mental,koch2018picture}.
For instance, when viewing a cup beside a laptop, we can quickly infer the cup’s approximate height using the known scale of the laptop and perspective cues such as relative position and occlusion.
This reasoning process does not rely on memorized numeric facts, but on contextual comparisons and geometric reasoning.
Such observations raise a key question: \textbf{do VLMs possess a comparable ability to infer metric scale from contextual references, rather than relying solely on rote memorization?}

\begin{figure}[h]
    \centering
    \includegraphics[width=1\linewidth]{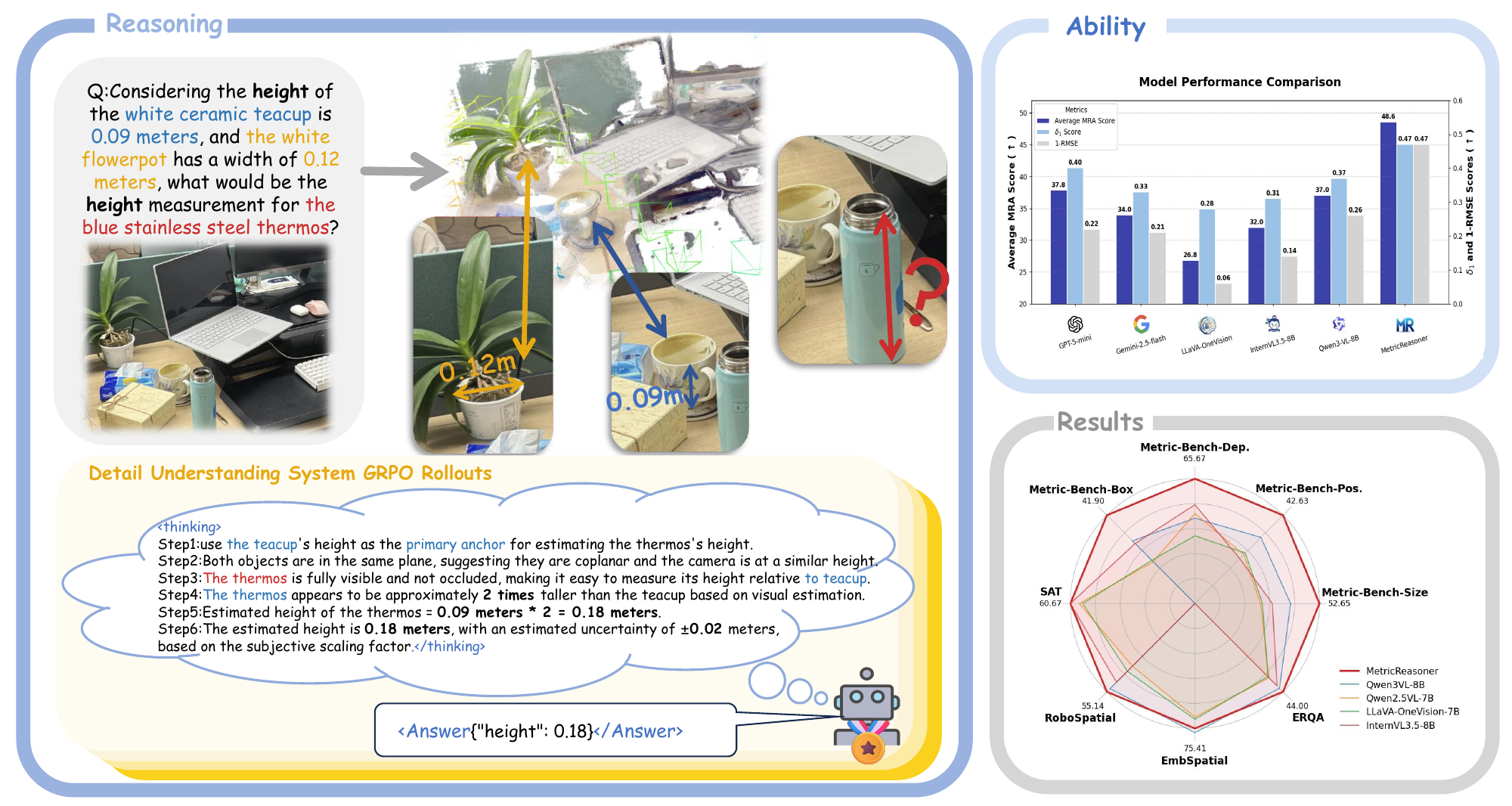}
    \caption{We propose Metric-Bench, a benchmark inspired by human perception of 3D space: given the true physical measurements of a few reference objects, the model learns to reason about relative positions and perspective relationships in 3D space, thereby inferring precise metric measurements of other objects in the scene. Furthermore, building upon this benchmark, we employ a spatial-guided reinforcement fine-tuning strategy, without compromising the model’s generalizable reasoning capability, and achieves state-of-the-art performance in metric scale perception, surpassing existing multimodal large language models.}
    \label{fig:main}
\end{figure}

To answer this question, this work introduces \textbf{Metric-Bench}, a new benchmark specifically designed to evaluate metric-scale reasoning in multimodal models.
Built upon the ScanNetV2 \cite{dai2017scannet} dataset, Metric-Bench includes 1340 carefully curated question–answer pairs that span five major spatial categories—size, position, depth, distance and 3D bounding box.
Each question provides multiple reference objects with a known metric and asks the model to infer another object’s absolute metric quantity.
Through this setup, Metric-Bench systematically tests whether a model can reason about real-world spatial metrics using in-image contextual cues rather than memorized priors.

Beyond evaluation, we further train the \textbf{MetricReasoner} model, using the proposed Reinforcement Fine-Tuning (RFT) framework which replaces supervised numerical regression with reward-based optimization over metric accuracy and structured outputs.
Instead of optimizing for answer accuracy alone, we design reward functions that encourage logically coherent, physically consistent reasoning steps grounded in reference metrics.
The supervision allows VLMs to learn structured spatial reasoning strategies—integrating geometric consistency, perspective cues, and contextual priors—thereby improving both interpretability and accuracy in metric reasoning tasks.

Our main contributions are as follows:
\begin{itemize}
    \item We introduce Metric-Bench, the first benchmark tailored to evaluate \textit{spatial metric reasoning} in multimodal large models—i.e., inferring absolute metric quantities from \emph{sparse} object–metric references—without requiring access to camera intrinsics.

    \item  We introduce a metric-guided RFT recipe tailored to reference-based spatial metric reasoning.
    By leveraging contextual reference metrics and Chain-of-Thought (CoT) reasoning, our method unlocks the latent spatial reasoning potential of existing vision–language models.

    \item Extensive experiments, including zero-shot and reinforcement fine-tuning (RFT) settings, demonstrate significant improvements in metric understanding while preserving general inference capabilities. We hope our analysis provides insights for advancing physically grounded and interpretable multimodal reasoning.

\end{itemize}

\section{Related Work}
\label{sec:related}

\subsection{Metric-Scale Reasoning}
Estimating metric depth and recovering spatial object geometric attributes from monocular images are long-standing challenges in computer vision, primarily due to the inherent scale ambiguity.
Traditional approaches address this by incorporating semantic priors \cite{zhang2025detect,wang2021fcos3d,wang2022detr3d}, leveraging the rich prior knowledge embedded in extensively pre-trained 2D foundation models to compensate for this scarcity, while Metric3D series\cite{yin2023metric,hu2024metric3dv2} utilize the camera intrinsics to normalize images and labels so that the model could effectively learn a unified metric scale of the world even when mixing with data from different types of cameras. To remove the need of camera intrinsics, recent works \cite{Bochkovskii2024,wang2025moge,wang2025moge2} designed specific architectures to address camera ambiguity by aligning with affine-invariant point cloud, along with scale optimization to calculate final metric scale. Despite their progress, these methods often overlook the human capacity to infer scale via in-context reference objects. 
In this work, we emulate this human-centric perceptive process, enabling models to resolve scale ambiguity by leveraging contextual scale priors rather than relying solely on rigid camera parameters.

\subsection{Spatial and Metric Benchmarks for VLMs}

Evaluating the spatial intelligence of Vision Language Models (VLMs) has gained significant traction. 
Recent benchmarks and methods~\cite{dongfang2025multimodal,jia2025omnispatial,wang2025site,ma20253dsrbench,yang2025thinking,song2025robospatial} have incorporated various aspects of metric reasoning into their evaluation protocols. 
For instance, 3DSR-Bench~\cite{ma20253dsrbench} and OmniSpatial~\cite{jia2025omnispatial} design image-based QA tasks that probe metric understanding, while VSI-Bench~\cite{yang2025thinking} extends this evaluation to video-based metric estimation scenarios.
Concurrent efforts aim at real-world applications. ERQA\cite{gemini2025robotics} systematically investigates trajectory reasoning, action estimation and task reasoning grounded in robotics scenarios, and RoboSpatial \cite{song2025robospatial} further assesses spatial awareness within the context of robotics and dynamic environments.
However, since inferring 3D metrics from 2D observations is inherently ill-posed, these benchmarks predominantly adopt multiple-choice or relative comparison, thus such qualitative assessments allow VLMs to bypass rigorous geometric reasoning. 
Existing benchmarks only support coarse relative judgments, while precise 3D metric reasoning remains underexplored. In contrast, our benchmark enables direct and accurate metric prediction by leveraging in-context visual cues and implicit object scale priors, which poses a more challenging and practical test for the spatial intelligence.

\subsection{Learning Paradigms for Spatial Intelligence}

A common approach to endow VLMs with spatial capabilities is to introduce dense pixel-level supervision or explicit numerical regression objectives~\cite{zong2025ground,liang2025pixelvla,munasinghe2025videoglamm,chen2024spatialvlm,cheng2025sr3d}. 
While effective on the targeted spatial tasks, such objective design can bias optimization toward local cues and task-specific outputs, which may undermine broader multimodal reasoning and sometimes manifests as transfer degradation or catastrophic forgetting.
To reduce this trade-off, recent works explore Chain-of-Thought (CoT) prompting and reinforcement learning to elicit intermediate reasoning and internalize multi-step logic via reward-driven learning~\cite{chen2025reasoning,ji2025enhancing}. 
Notably, RFT-based methods~\cite{liao2025improved,huang20253d,qi2025vln,jin2025spazer} have shown improvements in multi-step reasoning, but they typically employ outcome-level rewards or coarse spatial constraints, providing limited supervision over \emph{metric consistency} along the reasoning trajectory.
Different from these efforts, we use task-specific verifiable rewards to evaluate output validity and numerical proximity, while structured prompts encourage the model to use reference metrics as contextual evidence.
\section{Benchmark Construction Pipeline}
\label{sec:bench}

We introduce \textbf{Metric-Bench}, a benchmark specifically designed to evaluate spatial metric understanding capabilities of vision-language models, and primarily focuses on the absolute sizes and positions of instances in indoor scenes. 
Metric-Bench currently comprises $1340$ carefully curated question–answer pairs, each pair contains multiple real metrics of reference instances and requires the metric of a target object in the scene, which enables models to perform deep 3D spatial reasoning about metric scale based on relative positional relationships between instances in the projected image plane.

\begin{figure*}[t]
    \centering
    \includegraphics[width=1\linewidth]{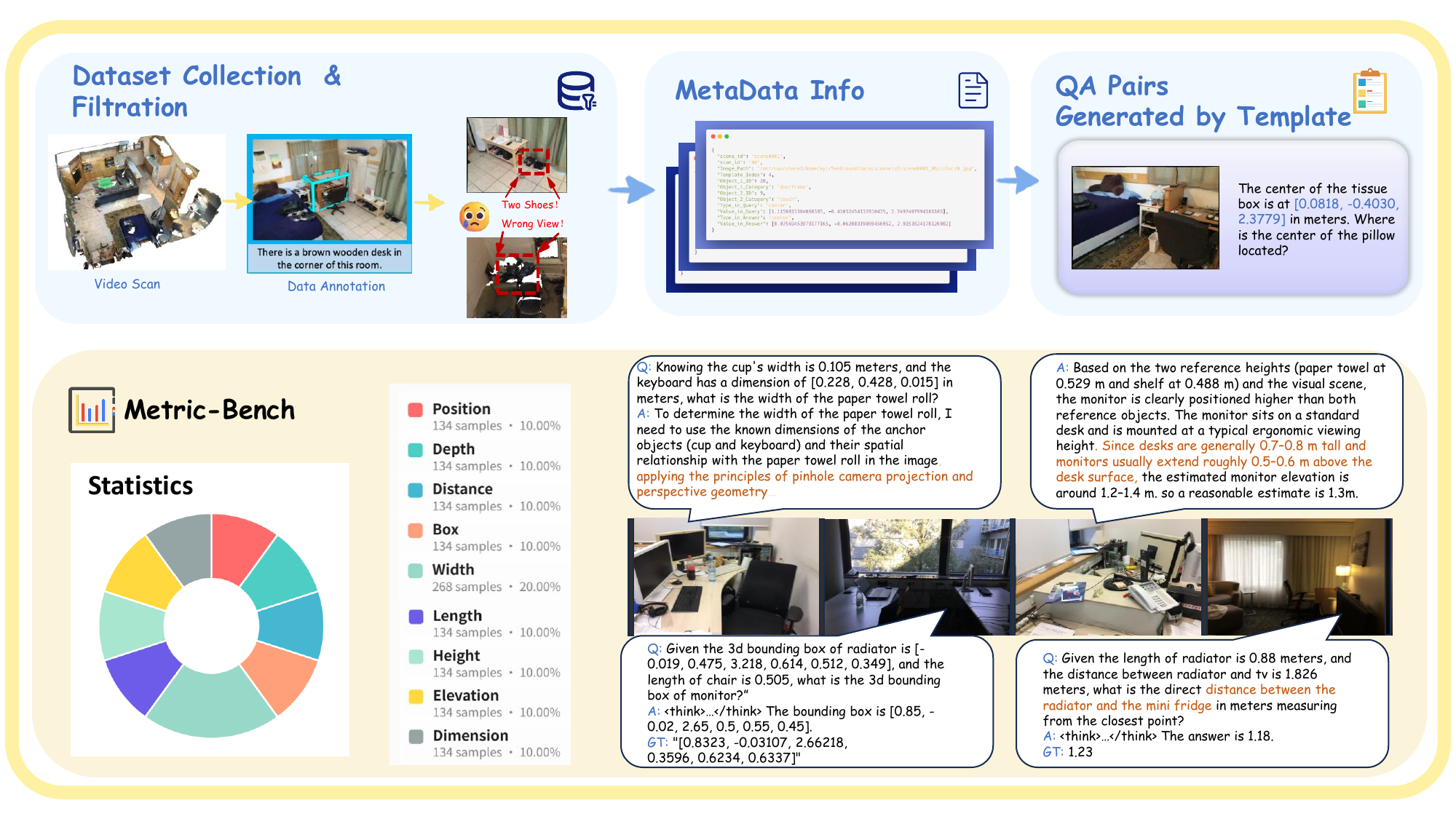}
    \caption{Data construction pipeline and statistics in Metric-Bench.}
    \label{fig:data}
\end{figure*}

\subsection{Data Collection}

Metric-Bench is constructed from ScanNetV2~\cite{dai2017scannet}, a real-world indoor RGB-D dataset that provides per-frame camera intrinsics/extrinsics and 3D instance annotations.
We focus on \emph{absolute metric} spatial reasoning grounded in natural ego-view observations.
Accordingly, spatial measurements in Metric-Bench are organized into five categories: \textit{size}, \textit{position}, \textit{depth}, \textit{distance}, and \textit{3D bounding box}.
The benchmark is formulated as question--answer pairs, where each question specifies multiple \emph{reference metrics} and asks the model to infer a \emph{target metric} under the same camera view.
The overall construction pipeline and dataset statistics are summarized in Fig.~\ref{fig:data}.

While ScanNetV2 annotates 3D bounding boxes in the world coordinate system, human perception and vision-based measurement are naturally performed in the \emph{ego} (camera) coordinate system and its projection onto the image plane.
To align the benchmark with this observation process and to make metric reasoning perspective-aware, we convert the original 3D annotations from the world coordinates to the camera coordinates using the provided camera pose.
In the camera coordinate system, $(x,y,z)$ denotes the object center location, and the length/width/height are defined along the object’s longest, shortest, and vertical axes, respectively.
This transformation standardizes metric definitions under a single view and ensures consistency with human-like size and distance estimation.

Metric-Bench is constructed in three steps:

\textbf{Step 1: Filtering images and annotations.}
ScanNetV2 scenes are captured as videos, where motion blur can significantly degrade both object visibility and measurement reliability.
We therefore remove blurry frames by computing the Variance of Laplacian\cite{pertuz2013analysis}, and retain only sufficiently sharp images.
For each remaining image, we project the 3D object centers and extents (in camera coordinates) onto the image plane using camera intrinsics, and keep only objects that are visible in the field of view.
This yields, for each selected image $i$, a set of valid 3D bounding boxes aligned with the corresponding ego-view observation.

\textbf{Step 2: Query generation.}
For each filtered image, we extract the ground-truth metric attributes from its valid 3D bounding boxes and instantiate predefined question--answer templates with object and metric identifiers from the metadata.
For example,
\emph{``If the \{object$_1$\} has a length of \{length$_1$\} meters, the \{object$_2$\} has a height of \{height$_2$\} meters, and the \{object$_3$\} has a width of \{width$_3$\} meters, what is the length of the \{object$_4$\}?''}
The answer is \{length$_4$\}.
To ensure diversity in object categories and spatial configurations, we require each image to contain at least four instances from distinct categories.

The templates cover three types:
(i) \textit{size-to-size inference}, which estimates a target object’s size from a reference object’s known size;
(ii) \textit{position/distance inference}, which deduces target position or distance from spatial cues of reference objects;
(iii) \textit{hybrid reasoning}, which jointly combines size and positional/distance cues.
By varying object metrics and spatial layouts, the resulting queries evaluate whether a model can robustly reason about absolute metric quantities under relative spatial context and perspective constraints.

\textbf{Step 3: Human verification.}
We apply a dual verification protocol combining an automatic VLM-based checker with manual inspection to remove queries with invisible targets, duplicate questions, or ambiguous cases (e.g., multiple plausible targets).
This step ensures accuracy, consistency, and minimal ambiguity of the benchmark annotations.

Following the above pipeline, we generate two isolated subsets: a 5K-sample training set (1,340 images) and a 1,340-sample test set (134 images).
Importantly, the two subsets are drawn from \emph{non-overlapping scenes} to prevent data leakage.
All samples are curated for camera-view metric measurement tasks under a unified, standardized construction procedure.

\section{Metric-guided Fine-tuning Strategy}

To address the current limitations of VLMs in accurate metric reasoning, we develop a sparse reference-based spatial model through a systematic enhancement approach, integrating the proposed Metric-Bench with a reinforcement learning(RL) fine-tuning strategy specifically designed for in-context spatial-metric understanding.
The fine-tuning strategy explicitly trains the model to reason about other objects in a scene relative to reference object metrics, enabling the model to learn a more unified and accurate representation of spatial metric relationships and thereby enabling robust metric-aware spatial reasoning in the absence of camera intrinsics.

To better elicit the inherent scene understanding and reasoning capabilities of VLMs, we design a specialized prompt and verifiable rewards for 3D metric reasoning. The prompt explicitly guides the model to generate a CoT by leveraging the given metrics of reference objects, thereby enabling step-by-step inference of scene-metric information. The verifiable rewards function contains three parts, instructing the model to learn the referring metric-guided thinking process and the metrics of target objects.

\textbf{Format Reward} To enforce structural compliance in model outputs, we introduce a format reward that evaluates whether the model encapsulates its reasoning process and final answer within the designated delimiters, \eg, \texttt{<thinking>\allowbreak...</thinking>\allowbreak<Final answer>...</answer>}.

\begin{equation}
    R_{format} = 
    \begin{cases} 
    1, & \text{if } \hat{y} \text{ matches format}, \\
    0, & \text{otherwise}.
    \end{cases}
\end{equation}
where $\hat{y}$ is the prediction of target metric value.

To enable the model to generate accurate metric predictions, we introduce two complementary numerical reward functions.

\textbf{Exponential Precision(EP) reward} The exponential precision reward continuously evaluates the prediction accuracy using an exponential decay over the absolute distance error:
\begin{equation}
    R_{EP}= e^{-|\hat{y} - y|},
\end{equation}
where $y$ is the ground-truth of the target metric.
Under the exponential precision reward scheme, the reward value diminishes exponentially with increasing absolute error—i.e., larger deviations from the ground-truth result in substantially lower rewards.

\textbf{Binned Proximity(BP) reward} Furthermore, a binned proximity reward discretizes the deviation between prediction and ground-truth into hierarchical intervals to provide coarse-grained supervision:
\begin{equation}
    R_{BP} = 
    \begin{cases} 
    1, & \text{if } 0 \le |\hat{y} - y| < 0.25, \\
    0.05, & \text{if } 0.25 \le |\hat{y} - y| < 0.5, \\
    0.01, & \text{if } 0.5 \le |\hat{y} - y| < 1, \\
    0, & \text{otherwise}.
    \end{cases}
\end{equation}

\textbf{Optimization} The final reward $R_{total}$ is computed as a sum of the format, precision, and proximity rewards:
\begin{equation}
    R_{total} = R_{format} + R_{EP} + R_{BP},
\end{equation}

The combined reward encourages the model to leverage contextual guidance to reason about spatial relationships, and produce metric-aware final outputs.

The reinforcement fine-tuning strategy builds on Group Relative Policy Optimization (GRPO) \cite{shao2024deepseekmath}. Specifically, for each query $q$, GRPO samples a group of outputs $\{o_1, o_2, \dots, o_G\}$ from the old policy $\pi_{\theta_{old}}$ and optimizes the following objective:
\begin{small}
\begin{equation}
\begin{aligned}
\begin{split}
    &\mathcal{J}_{GRPO}(\theta) = 
     \mathbb{E}_{q \sim P(Q), \{o_i\}_{i=1}^G \sim \pi_{\theta_{old}}(O|q)} \\ 
    & \left[ \frac{1}{G} \sum_{i=1}^G \left( \min \left( \rho_i A_i, \text{clip}(\rho_i, 1-\epsilon, 1+\epsilon) A_i \right)  - \beta D_{KL}(\pi_\theta || \pi_{ref}) \right) \right],
\end{split}
\end{aligned}
\end{equation}
\end{small}
where $\rho_i = \frac{\pi_\theta(o_i|q)}{\pi_{\theta_{old}}(o_i|q)}$ is the probability ratio, and the advantage $A_i$ is computed by normalizing the total rewards within the group:
\begin{equation}
    A_i = \frac{R_{total,i} - \text{mean}(\{R_{total,1}, \dots, R_{total,G}\})}{\text{std}(\{R_{total,1}, \dots, R_{total,G}\}) + \delta}.
\end{equation}

\section{Experiments}
\label{sec:experiments}

\begin{table*}[t]
    \caption{The metric estimation performance of public and our fine-tuned models on Metric-Bench. All public models are evaluated under zero-shot settings. Proprietary models and human performance are evaluated using a mini dataset, which contains 400 queries randomly sampled from the bench.  Pos., Dep., Dis. are abbreviations for position, depth and distance. Length, width, height, elevation and dimension are all included in Size. The unit of RMSE is meters.
    }
    \centering
    \small
    \begin{tabular}{r|cccccc|c|c}
        \toprule
         & \multicolumn{6}{c|}{MRA$\uparrow$} &\multirow{2}{*}{RMSE$\downarrow$}  & \multirow{2}{*}{${\delta_1}\uparrow$} \\
         & Avg. & Size  & Pos. & Dep. & Dis. & Box &  &  \\
        \midrule
        Human & 76.83 & 81.51 & 77.76 &76.33 & 79.35 & 82.09 & 0.2671 & 0.7920  \\
        \midrule
        \rowcolor{gray!10} 
        Proprietary Models (API)-Mini & &  & & &  & &  &  \\
        Gemini-2.5-flash & 33.95 & 34.74 & 31.96 & 61.64 & 18.87 & 29.09 & 0.7886 &  0.3254 \\
        GPT-5-mini & 37.84 & 38.47 & 33.06 & 58.88 & \textbf{22.73} & \underline{33.40} & 0.7842 & 0.4014 \\
        \midrule
        \rowcolor{gray!10} 
        Open-source Models  & &  &  & & & &  & \\
        LLaVA-OneVision-Qwen2-0.5B & 20.65 & 23.77 & 23.25 & 7.99 & 14.05  & 25.49 & 1.2750 &  0.2105  \\
        LLaVA-OneVision-Qwen2-7B & 26.81 & 28.55 & 24.38 & 35.62 & 16.15 & 20.52 &  0.9367 & 0.2763 \\
        InternVL3.5-1B & 23.64  & 20.87 & 32.30 & 41.71& 15.13 & 23.33 & 1.0776 & 0.2581  \\     
        InternVL3.5-4B & 33.72 & 35.96 & 26.50 & 49.93 & 19.48& 24.63& 0.8256 & 0.3298  \\
        InternVL3.5-8B & 32.00 & 32.72 & 21.71 & 51.97 & 21.46& 28.47& 0.8637 & 0.3116  \\
        InternVL3.5-14B & 31.49 & 37.85 & 16.75& 24.56 & 17.98 & 28.31& 0.9400 & 0.3135  \\
        Qwen2.5-VL-7B & 27.58 & 27.75 & 23.66 & 47.34 & 16.71 & 19.84 & 0.7404 &  0.2577  \\ 
        Qwen2.5-VL-32B & 32.71 & 33.82 & 24.93 & 50.98 & 20.45 & 28.07 & 0.8647 & 0.3196  \\ 
        Qwen3-VL-8B &  37.04 & 40.44& 31.93 & 44.91 & 21.14 & 29.73 & 0.7394 & 0.3695  \\
        Qwen3-VL-32B &  39.24 & 42.11 & \underline{34.74} & 49.80 & \underline{22.28} & 31.83 & 0.7331 & 0.3836  \\
        \midrule
        \rowcolor{gray!10} 
        Spatial-specialized Models  & &  &  & & & &  & \\
        VST-7B-RL & 38.11 & 40.76& 27.63& \underline{55.43} &17.44 &26.52 & \underline{0.7210} & 0.4039 \\
        VST-7B-SFT & \underline{39.61}  & \underline{42.23} &28.58 &53.84 & 19.63&26.86 &  0.7463 & \underline{0.4058} \\
        SpaceR-7B  & 28.97  & 27.66 &23.97 & 52.49&22.15 &24.11 &  0.8983 & 0.2947 \\
        \midrule
        \midrule
        MetricReasoner & \textbf{48.59} & \textbf{52.65} & \textbf{42.63} & \textbf{65.67}&19.81 & \textbf{41.90} &\textbf{0.5323} & \textbf{0.4705} \\
        \bottomrule
    \end{tabular}
    \label{tab:main}
\end{table*}
\subsection{Experiment Setup}
\textbf{Baselines and Metrics} We comprehensively evaluate 15 VLMs that include a diverse set of models spanning
different architectures and parameter scales under zero-shot settings: (1) Proprietary models: GPT-5-mini \cite{2025gpt5} and Gemini-2.5-Flash \cite{comanici2025gemini}. (2) Open-source
models: InternVL3.5 series \cite{wang2025internvl3_5}, 
Qwen-VL series including Qwen2.5-VL \cite{Qwen2.5-VL} and Qwen3-VL \cite{Qwen3-VL}, and LLaVA-OneVision \cite{li2024llava} models. (3)Spatial-specialized model: VST\cite{yang2025visual} models and SpaceR\cite{ouyang2025spacer}.
Human evaluation results on Metric-Bench are also reported for benchmarking and comparative reference.

For evaluation metrics, we follow VSI-Bench \cite{yang2025thinking} and use MRA (Mean Relative Accuracy) as our main metric. Besides, RMSE (Root Mean Squared Error) and $\delta_1$ are also considered for a more intuitive performance comparison. 
For each prediction and ground-truth pair, the MRA is defined as
\begin{equation}
    MRA = \frac{1}{n} \sum_{i=1}^{n} \mathbb{1} \left( \frac{|\hat{y} - y|}{y} \leq 1 - \theta_i \right),
\end{equation}
where $\theta_i = \{0.5, 0.55, . . . , 0.95\}$, and $n=10$.

The RMSE is defined following the depth estimation task, \ie,
\begin{equation}
    RMSE = \sqrt{ \frac{1}{N} \sum_{i=1}^{N} \left( \hat{y}_i - y_i \right)^2 }
\end{equation}
where N is the number of all test queries.
The $\delta_1$ metric is defined as,
\begin{equation}
    \delta_1 = \frac{1}{N} \sum_{i=1}^{N} \mathbb{1} \left( \max\left( \frac{y_i}{\hat{y}_i}, \frac{\hat{y}_i}{y_i} \right) < 1.25 \right).
\end{equation}

\textbf{Implementation Details} To evaluate the relative efficacy of different fine-tuning paradigms, we apply both supervised fine-tuning (SFT) and reinforcement fine-tuning (RFT) to the Qwen3-VL-8B base model. MetricReasoner refers to the model that employs the RFT strategy. Both RFT and SFT Models are trained for 3 epochs to ensure full convergence and optimal performance. 
All experiments are implemented on 4 H100 GPUs with batch size of 128, group size of 32, and learning rate of $2\times 10^{-6}$. Training for 3 epochs takes approximately 12 hours. We perform the scene-level split before QA generation. The training split used both in RFT and SFT, is constructed from ScanNetV2 training scenes, while Metric-Bench test samples are drawn from non-overlapping test scenes. Therefore, no scene, image, or object instance from Metric-Bench is used during RFT.

\begin{figure}
    \centering
    \includegraphics[width=0.8\linewidth]{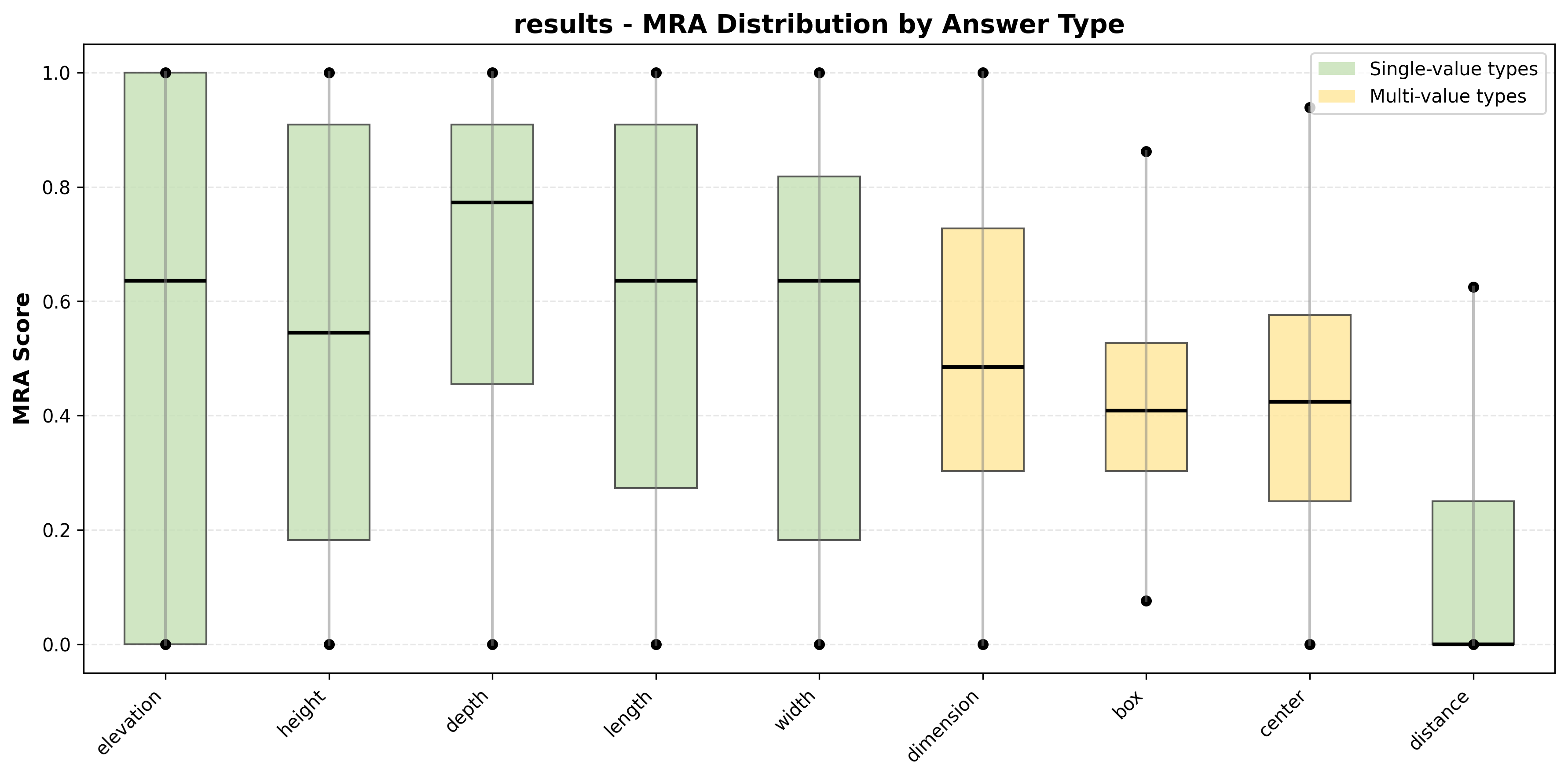}
    \caption{Performance of MetricReasoner across all categories in Metric-Bench.}
    \label{fig:rft}
\end{figure}

\subsection{Main Results}

Table \ref{tab:main} presents a comprehensive comparison of different models across multiple metrics on Metric-Bench. The Size denotes the average of MRA in dimensions, \ie, width, length, height, elevation and dimension.
Overall, proprietary models outperform most open-source counterparts, while spatial-specialized achieves the highest MRA score of 39.61, notably surpassing Gemini-2.5-flash (33.95), GPT-5-mini (37.84) and also exhibits better performance in both RMSE and ${\delta_1}$.
For open-source models, the InternVL series demonstrates relatively stable results, where InternVL3.5-4B shows competitive performance among all InternVL series. 
Meanwhile, Qwen3-VL-32B delivers the best results among the evaluated open-source general-purpose models, reaching 39.24 in MRA and 0.3836 in ${\delta_1}$. 
In comparison, our proposed MetricReasoner significantly surpasses all baseline models. MetricReasoner, representing metric-guided RFT strategies, achieves the best MRA scores of 48.59, with ${\delta_1}$ increasing to 0.4705, while substantially reducing RMSE to 0.5323. These results clearly demonstrate the superiority of our method in multi-dimensional spatial metric understanding and depth consistency modeling. Besides, it is worth noting that among all categories, most models perform better on depth estimation than on other metric items, with distance estimation yielding the poorest performance, which leaves challenges for further improvement on these metrics.

The detailed performance of our MetricReasoner is illustrated in Fig. \ref{fig:rft}, which indicates that models generally perform better on single-value categories, with more consistent and reliable MRA scores. Multi-value categories present a greater challenge, reflecting a higher degree of variability and lower prediction accuracy. Among the single-value categories, the score range for the distance metric is comparable to that of other attributes, but its median is lower, reflecting the model's performance shortcomings on this category.

\subsection{Ablation Study}

\textbf{Ablation of the number of referenced metrics}
Table.\ref{tab:ft} studies how the number of referenced metrics affects MetricReasoner. Without any reference, the model lacks an explicit calibration anchor and yields relatively low accuracy (MRA=47.56) and higher error (RMSE=0.5540). Introducing a small set of references consistently improves performance: $N_{ref.}$=2 increases $\delta_1$ to 0.4852 and reduces RMSE to 0.5392, while $N_{ref.}$=4 achieves the best overall trade-off, giving the highest MRA (48.59) and the lowest RMSE (0.5323). However, further increasing the references to 5 leads to a clear degradation across all metrics, 
suggesting diminishing returns and potential information overload that hinder stable calibration. Overall, a moderate number of referenced metrics (e.g., 4) is optimal, and more references are not necessarily better.

\textbf{Ablation of fine-tuning rewards}
Table~\ref{tab:reward} ablates the two accuracy rewards, EP and BP. Removing both rewards severely hurts performance, indicating that accuracy-oriented reinforcement is crucial for reliable metric prediction. Enabling either reward alone yields substantial gains: EP-only improves MRA to 45.95 and reduces RMSE to 0.5798, while BP-only further boosts MRA to 48.05 and $\delta_1$ to 0.4537 with a comparable RMSE (0.5587), suggesting binned proximity reward provides a stronger supervision signal for correct metric calibration. Combining EP and BP leads to additive improvements. Overall, BP appears to be the dominant contributor, and a careful reward balancing strategy may be required to fully benefit from the combination.

\begin{table}[t]
\centering
\begin{minipage}[t]{0.48\textwidth}
    \centering
    \captionof{table}{The performance of various amounts of referenced metrics.}
    \label{tab:ft}
    \begin{tabular}{cccc}
        \toprule
           $N_{ref.}$  & MRA$\uparrow$ & RMSE$\downarrow$ & ${\delta_1}\uparrow$ \\
        \midrule
         0 & 47.56 & 0.5540 & 0.4586 \\
         2 & 48.26 & 0.5392 & \textbf{0.4852} \\
         4 & \textbf{48.59} & \textbf{0.5323} & 0.4705 \\
         5 & 46.85 & 0.5604 & 0.4382 \\
         \bottomrule
    \end{tabular}
\end{minipage}
\hfill
\begin{minipage}[t]{0.48\textwidth}
    \centering
    \captionof{table}{Ablation on two accuracy rewards BP and EP.}
    \label{tab:reward}
    \begin{tabular}{cccccc}
        \toprule
        EP & BP & MRA$\uparrow$ & RMSE$\downarrow$ & $\delta_1 \uparrow$ \\
        \midrule
        $\times$ & $\times$ & 12.38 & 1.0515 & 0.0862 \\
        $\checkmark$ & $\times$ & 45.95 & 0.5798 & 0.4281 \\
        $\times$ & $\checkmark$ & 48.05 & 0.5587 & 0.4537 \\
        $\checkmark$ & $\checkmark$ & \textbf{48.59} & \textbf{0.5323} & \textbf{0.4705} \\
        \bottomrule
    \end{tabular}
\end{minipage}
\end{table}

\subsection{Performance on Out-of-domain Spatial Benchmarks}
\label{OOD}
To verify the preservation of the spatial reasoning capability of MetricReasoner, we evaluate our model on several spatial understanding benchmarks.
Table~\ref{tab:merged_general} evaluates downstream embodied benchmarks on ERQA\cite{gemini2025robotics} and RoboSpatial\cite{song2025robospatial}. Overall, MetricReasoner consistently delivers the strongest transfer performance, outperforming both the general VLM baseline and spatial-specialized models. On ERQA, MetricReasoner achieves the best accuracy (44.00), surpassing Qwen3-VL-8B (42.25) and the spatial-focused VST-7B-RL (40.25). On RoboSpatial, MetricReasoner further sets a new high in overall accuracy (55.14) and improves key sub-dimensions, notably Compatibility (56.19) and Configuration (83.74). In contrast, spatial-specialized models exhibit weaker embodied transfer, with VST-7B-RL trailing substantially in RoboSpatial Acc. (42.29) and particularly struggling on Context (1.63). These results indicate that MetricReasoner learns spatial reasoning signals that generalize effectively to embodied decision-centric evaluations, yielding robust gains beyond both the base model and prior spatial-specialized alternatives.


\subsection{Performance on General Purpose Benchmarks}
Table~\ref{tab:merged_general} also reports results of three models on three general purpose benchmarks \ie, CountBench\cite{paiss2023countclip}, V$\star$ Bench\cite{vstar}, and BLINK\cite{fu2024blink}). In general-purpose evaluation, VST-7B-RL lags behind the strong general VLM baseline Qwen3-VL-8B on all three benchmarks (e.g., 59.16\% vs. 61.78\% on V$\star$ Bench and 43.32\% vs. 58.61\% on BLINK), suggesting that VST-7B-RL exhibits limited performance on broad visual–language capabilities. In contrast, MetricReasoner not only maintains general understanding ability but also improves it: it achieves the best performance across the board, reaching 0.920 on CountBench, 68.59\% on V$\star$ Bench, and 81.82\% on BLINK. These results indicate that our method enhances spatial reasoning without sacrificing generalization, and even delivers consistent gains over the general baseline, demonstrating robust transfer beyond the targeted spatial benchmarks.

\begin{table*}[t]
\centering
\small
\caption{General performance of spatial-specialized model VST-7B-RL, our MetricReasoner and corresponding base model Qwen3-VL-8B on ERQA, RoboSpatial, and general benchmarks, including CountBench, V$\star$ Bench and BLINK. }
\label{tab:merged_general}
\begin{tabular}{l c c c c c c c}
\toprule
Models & ERQA $\uparrow$ & RoboSpatial$\uparrow$  & CountBench$\uparrow$ & V$\star$ Bench$\uparrow$ & BLINK$\uparrow$ \\
\midrule
Qwen3-VL-8B     & 42.25 & 53.14  & 0.917 & 61.78\% & 58.61\% \\
VST-7B-RL       & 40.25 & 42.29  & 0.891 & 59.16\% & 43.32\% \\
MetricReasoner  & \textbf{44.00} & \textbf{55.14}  & \textbf{0.920} & \textbf{68.59\%} & \textbf{81.82\%} \\
\bottomrule
\end{tabular}
\end{table*}

\subsection{Comparison between SFT and RFT strategies}

\begin{table}[t]
\centering
\caption{Results on RoboSpatial, CountBench, V$\star$ Bench, and BLINK of different fine-tuning strategies. RL w/ COT\&SFT denotes SFT on CoT-annotated data followed by reinforcement fine-tuning.}
\small
\begin{tabular}{lcccc}
\toprule
Strategy & RoboSpatial & CountBench &  V$\star$ Bench & BLINK \\
\midrule
RL(MetricReasoner)      & \textbf{55.14} & \textbf{0.9204} & \textbf{68.59\%} & \textbf{81.82\%} \\
RL w/ COT\&SFT & 45.14 & 0.9190 & 62.83\% & 34.53\% \\
SFT     & 40.86 & 0.9093 & 63.35\% & 51.26\% \\
\bottomrule
\end{tabular}
\label{tab:sft}
\end{table}

Table \ref{tab:sft} shows the influence of different fine-tuning strategies on the Metric-Bench. RL w/ COT\&SFT represents a two-stage fine-tuning. On stage1, we use the CoT data generated by Gemini-2.5\cite{comanici2025gemini} to fine-tune the base model, then our RFT strategy is employed on stage2. The results indicate that training strategies critically impact both spatial reasoning and general benchmark transfer: RL fine-tuning only(MetricReasoner) achieves the best overall performance, suggesting that reward-driven optimization yields spatially grounded behaviors that generalize well. In contrast, the RL w/ COT\&SFT fine-tuning degrades RoboSpatial to 45.14 and V$\star$ to 62.83\%, implying that CoT-style supervision may introduce patterns misaligned with the downstream reward objective and is not fully corrected by subsequent RFT. Pure SFT with CoT data performs worst on RoboSpatial (40.86) and trails RL on all general benchmarks, suggesting that supervised fine-tuning alone is less effective at inducing robust, transferable spatial reasoning than direct RL optimization.

\subsection{Qualitative results of MetricReasoner}

\begin{figure*}[t]
    \centering
    \includegraphics[width=1\linewidth]{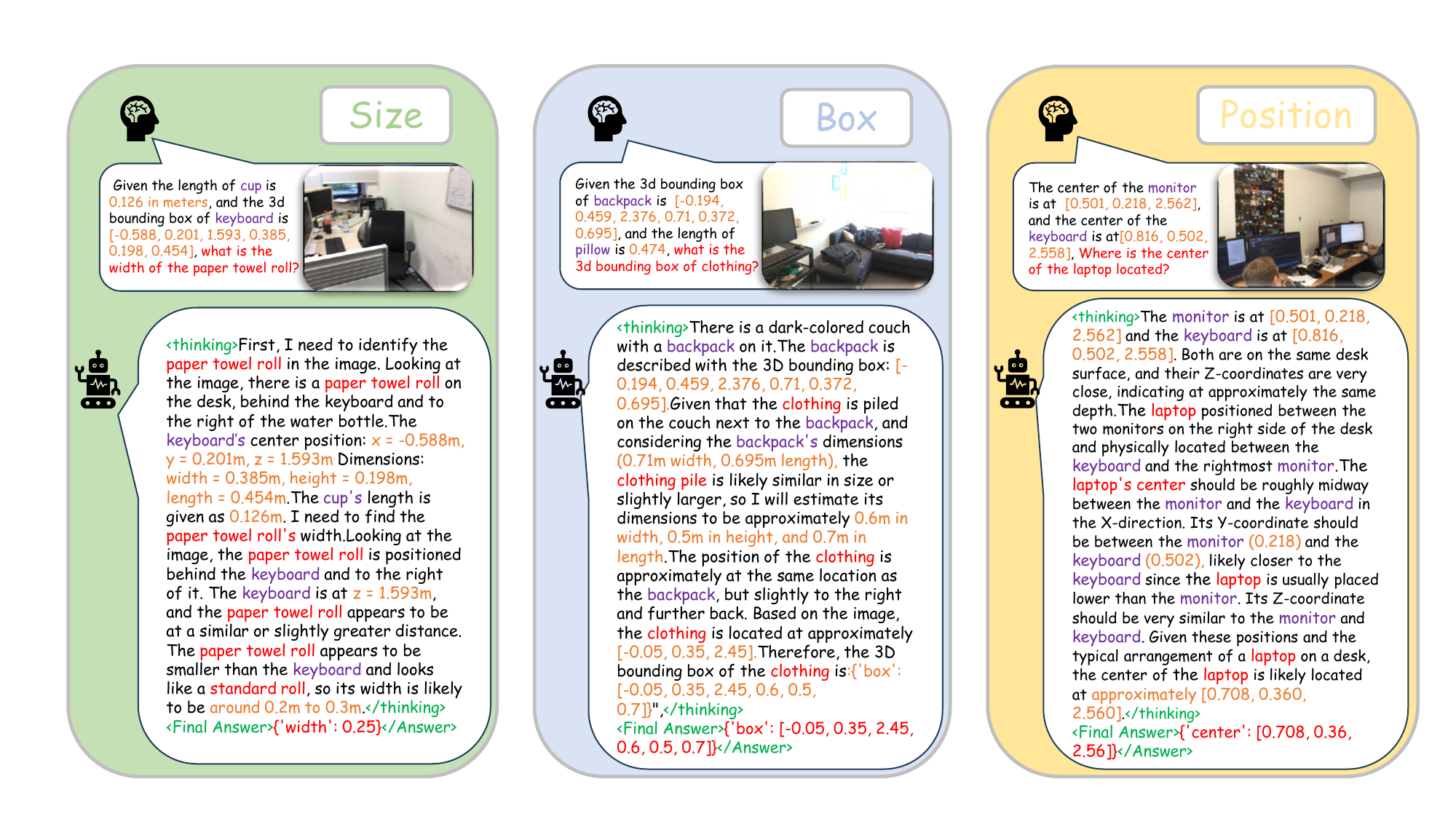}
    \caption{Qualitative results of MetricReasoner.}
    \label{fig:qualitative results of metric-bench}
\end{figure*}

To provide an in-depth understanding of the spatial reasoning capabilities of our MetricReasoner model, we present qualitative results that showcase the model's CoT process in inferring metric quantities from in-image contextual cues. Fig. \ref{fig:qualitative results of metric-bench} illustrates three representative cases across different metric categories: Size, Box, and Position. These examples demonstrate how the model integrates visual evidence with metric anchors and perspective cues to arrive at a physically consistent metric prediction, moving beyond mere semantic recall.

\textbf{Size Estimation.} The visualization refers to size estimation. The model is asked to estimate the width of a paper towel roll, using nearby objects such as the keyboard and cup as contextual references. The model's CoT first identifies the relative position of the target object and referring objects, as the paper towel roll located on the desk, behind the keyboard and to the right of the water bottle. It then establishes metric cues from the scene, including the keyboard's 3D position and dimensions, as well as the cup's given length ($0.126$ m). Based on the relative visual scale, the model observes that the paper towel roll is smaller than the keyboard but comparable to a standard desktop object. The reasoning process is reasonable: it first localizes the target object, then anchors the estimation with nearby objects of known size, and finally combines relative comparison with object-size prior knowledge to infer that the paper towel roll's width is likely around $0.2 \sim 0.3$ m. This demonstrates the model's ability to leverage both in-image metric references and commonsense object priors for size estimation.

\textbf{3D Bounding Box Inference.} The second case illustrates the model's ability to infer the 3D bounding box of an unseen target object from nearby metric references. The target is a pile of clothing placed on a dark-colored couch, next to a backpack with a known 3D bounding box $[-0.194, 0.459, 2.376, 0.71, 0.372, 0.695]$. The model first localizes the clothing pile relative to the backpack, then uses the backpack's dimensions as a scale anchor to estimate the target size. Since the clothing pile is visually close to the backpack and appears slightly more spread out, the model predicts its dimensions as approximately $0.6$ m in width, $0.5$ m in height, and $0.7$ m in length. It further infers that the clothing pile is located slightly to the right and farther back from the backpack, resulting in the predicted box $[-0.05, 0.35, 2.45, 0.6, 0.5, 0.7]$. This example shows that the model can combine spatial relations, nearby 3D anchors, and object-level priors to produce a physically plausible 3D box estimation.

\textbf{3D Position Inference.} The final example (Position) asks for the center position of a laptop, referencing the known centers of a monitor and a keyboard. The model leverages these two anchors to establish the local desk layout and infer the laptop's position within the same $3$D coordinate frame. The CoT first observes that the monitor and keyboard have very similar $Z$-coordinates, indicating that they are placed at approximately the same depth on the desk surface. It then translates the $2$D spatial relationship of the laptop—located between the keyboard and the rightmost monitor—into $3$D coordinates by interpolating its $X$- and $Y$-positions between the two anchors while keeping a similar $Z$ value. This leads to the estimated laptop center $[0.708, 0.360, 2.560]$. This example demonstrates the model's ability to combine co-planarity, relative layout, and metric anchors to infer plausible 3D object positions from image observations.

These qualitative results validate our core hypothesis: MetricReasoner, by imposing structured reasoning guidance, encourages VLMs to transition from prior memorization to a structured, contextual metric reasoning strategy. The resulting CoT is both interpretable and physically grounded, making the model’s spatial inferences more reliable.

\section{Conclusion}
\label{sec:conclusion}

Inspired by how humans perceive 3D space through relative positioning, we introduce Metric-Bench, a benchmark designed to evaluate vision-language models' ability to infer the absolute spatial metric of a scene from a single image, given only sparse object-metric references. This setup eliminates reliance on camera intrinsics, thereby enabling robust adaptation across diverse real-world scenarios and generic imaging configurations.
Our analysis reveals that current VLMs excel notably in estimating metric depth compared to other metric dimensions (\eg, width or height), yet exhibit pronounced weaknesses in predicting metric distances between objects. 
On Metric-Bench, we demonstrate that implicitly supervising Chain-of-Thought (CoT) generation via a verifiable reward function substantially enhances spatial-metric reasoning. 
Our approach not only surpasses existing large proprietary models in metric estimation accuracy but also preserves strong generalization on downstream spatial understanding and manipulation tasks, without compromising the general reasoning capabilities.  

Future avenues of improvement include in-context absolute metric learning from video inputs and task-specific fine-tuning strategies tailored to downstream spatial reasoning applications.

\section{Acknowledgment}
This work was supported by the National Natural Science Foundation of China (No. 62576315, No. 62506338)

\bibliographystyle{splncs04}
\bibliography{main}

\clearpage
\clearpage
In this supplementary material, we provide the design of prompt in Sec. \ref{prompt}, our query template in Sec. \ref{sec:template}, detailed answer format in Sec. \ref{format}, 
and qualitative results of MetricReasoner on Metric-Bench and downstream manipulation tasks in Sec. \ref{resultMB} and Sec. \ref{result_mani}.
\section{Prompt Design}\label{prompt}

The prompt consists of three key components:
(1) \textbf{3D Fundamentals}, foundational concepts about the pinhole imaging mechanism, the definition of camera coordinate system and the format of 3D bounding box. 
(2) \textbf{Reasoning Guidance}, clear instructions on how to derive metric clues from visual and textual inputs, leveraging reference objects of known metric sizes. 
(3) \textbf{Output Specification}, a standardized format for structured, interpretable responses.
This structured design ensures consistent, traceable, and physically grounded metric reasoning.

\begin{promptbox}[Prompt]{gray}
    You are given an image along with a textual description that includes 3D metric information about certain anchor objects in the scene, represented in the camera coordinate system. Your task is to reason about the 3D spatial layout and infer the metric property (e.g., height) of a specified target object based on visual cues, spatial relations, and provided anchor metric measurement.
Prerequisite knowledge:

    1. **The Camera Coordinate System**: Recall that the camera coordinate system has its origin at the optical center, with:
    
       - **Z-axis** pointing in the imaging direction (forward),
       
       - **X-axis** parallel to the image horizontal (positive right),
       
       - **Y-axis** parallel to the image vertical (positive downward).
       
       All positions and dimensions are in meters.
       
    2. **Anchor box Information**: If the 3D bounding box of the anchor object (e.g., a book) is given, the format is `[x, y, z, width, height, length]', where: 
    `[x, y, z]' is the center position in camera coordinates, `[width, height, length]' correspond to the shorter horizontal dimension, vertical dimension, and longer horizontal dimension, respectively.
       
The question that need to be reasoned:

Please provide your response in two parts: **Thinking Process** and **Final Answer**.

3.**Thinking Process**: Your analysis where you explain your reasoning step-by-step, which includes but is not limited to the following aspects:

    a. **Define the Reference Anchor and Its Known Physical Quantity**  

    b. **Establish Spatial Pose and Contextual Layout of the Anchor**  

    c. **Identify and Locate the Target Object in Image Space**  

    d. **Construct Spatial Relationship Graph (with Optional Intermediates)**  
\end{promptbox}
\begin{promptbox}[Prompt]{gray}

    e. **Estimate Relative Quantity Offset or Scaling Ratio**  

    f. **Derive Absolute Quantity via Explicit Calculation Chain**  

    g. **Document Assumptions, Limitations, and Uncertainty Bounds**  

4. **Final Answer**: Your final numerical estimate for the target object’s physical quantity (e.g., height in meters), wrapped in `\texttt{<Final answer>}' tags and formatted as a JSON object with the appropriate key.

Example of the required output format which you should follow:
\texttt{<thinking>}Step-by-step reasoning following the seven aspects above, grounded in visual analysis and explicit calculation.\texttt{</thinking>} Please output the result in JSON format: \texttt{<Final answer>{`length': <value>}</answer>}.
\end{promptbox}

\section{Query Template}\label{sec:template}
To systematically evaluate a model’s capability in metric reasoning and spatial understanding within 3D scenes, we design a set of structured query templates for automatically generating question-answer samples that cover diverse spatial attributes. Specifically, these templates are organized into three categories. The first category, Dimensions, focuses on assessing the model’s ability to understand and infer geometric properties such as object length, width, height, and 3D bounding boxes. The second category, Position and Distance, examines spatial relationships including object centers in the camera coordinate system or ground reference frame, as well as depth, elevation, and relative distance. The third category, Complex and Mixed Queries, further combines geometric dimensions, positional relations, and local constraints to construct more challenging problems that require multi-step reasoning. By instantiating these templates with variables such as object identifiers, size attributes, center coordinates, bounding boxes, and pairwise relations, we can generate large-scale query samples with unified semantic structure, controllable difficulty, and broad coverage of reasoning patterns. This provides a fine-grained benchmark for evaluating a model’s spatial scale perception, geometric consistency modeling, and compositional reasoning ability.

\begin{promptbox}[Templates]{gray}

    \textbf{Dimensions}\\
    1. If the ${object_0}$ has a length of ${length_0}$ meters, the ${object_1}$ has a width of ${width_1}$ meters, the ${object_2}$ has a height of ${height_2}$ meters, what is the length of the ${object_3}$? \\
    2. Knowing the ${object_0}$'s width is ${width_0}$ meters and the ${object_1}$ has a dimension of ${dimension_1}$ meters, while the ${object_2}$ has a length of ${length_2}$ meters, what is the width of the ${object_3}$?
    \end{promptbox}

\begin{promptbox}[Templates]{gray}
    
    3. Considering the height of the ${object_0}$ is ${height_0}$ meters and the ${object_1}$ has a width of ${width_1}$ meters, the ${object_2}$ is ${depth_2}$ meters from the camera, what would be the height measurement for the ${object_3}$? \\
    4. If the ${object_0}$ has a dimension of ${dimension_0}$ in meters, the 3d bounding box of ${object_1}$ is ${box_1}$, the ${object_2}$ has a width of ${width_2}$ meters, what would be a corresponding dimension for the ${object_3}$? 

    \textbf{Position and Distance}\\
    5. The center of the ${object_0}$ is at ${center_0}$ in meters in camera coordinate system, the center of the ${object_1}$ is at ${center_1}$, the ${object_2}$ has a bounding box ${box_2}$, Where is the center of the ${object_3}$ located? \\
    6. If the ${object_0}$ is ${depth_0}$ meters away from the camera and the ${object_1}$ is ${depth_1}$ meters away, while the ${object_2}$ is positioned ${elevation_2}$ meters above the ground, how far is the ${object_3}$ from the same viewpoint? \\
    7. The ${object_0}$ is positioned ${elevation_0}$ meters above the ground and the ${object_1}$ is positioned ${elevation_1}$ meters above the ground; the ${object_2}$ has a length of ${length_2}$ meters, What is the ${object_3}$'s elevation from the ground? \\
    \textbf{Complex and Mixed Queries}\\
    8. Given the length of ${object_0}$ is ${length_0}$ meters, the distance between the ${object_0}$ and the ${object_1}$ is ${distance_1}$ meters, the ${object_2}$ has a height of ${height_2}$ meters, what is the direct distance between the ${object_2}$ and the ${object_3}$ in meters measuring from the closest point?
    
    9. Given the 3d bounding box of ${object_0}$ is ${box_0}$, the length of ${object_1}$ is ${length_1}$, the width of ${object_2}$ is ${width_2}$, what is the 3d bounding box of ${object_3}$? \\
    10. Given the length of ${object_0}$ is ${length_0}$ in meters and the 3d bounding box of ${object_1}$ is ${box_1}$, the ${object_2}$ has a height of ${height_2}$ meters, what is the width of the ${object_3}$?
\end{promptbox}

\section{Answer Format}\label{format}
For format of answers, length, width, height, depth, elevation and distance are single-value output, while dimension, center and box output a list that contains dimensions [width, height, length], coordinates [x, y, z] and both. The distance is defined by the shortest distance between objects and height is the object height for itself, while elevation represents the altitude above the ground. 

\section{Performance on Q-Spatial++}\label{Qspatial}
Metric-Bench differs from Q-Spatial Bench\cite{liao2024qspatial} in both task formulation and reference-based metric grounding. Q-Spatial Bench evaluates direct single-image quantitative estimation and uses SpatialPrompt to encourage models to find reference objects, but these references are not explicitly provided with known metrics, so monocular scale ambiguity remains. In contrast, Metric-Bench provides sparse reference-object measurements as in-context anchors, requiring models to perform reference-grounded scale calibration, and covers richer outputs beyond distance and size.
In addition, we evaluate MetricReasoner on Q-Spatial++ bench in a fully zero-shot way. This bench is captured from real scenes and is out-of-distribution for our model, MetricReasoner achieves 64.77 RMSE, compared with 143.54 for Qwen3-VL-8B and 72.58 for Gemini-2.5-Flash(cm).

\section{Qualitative results on Metric-Bench}\label{resultMB}

\begin{figure*}
    \centering
    \includegraphics[width=1\linewidth]{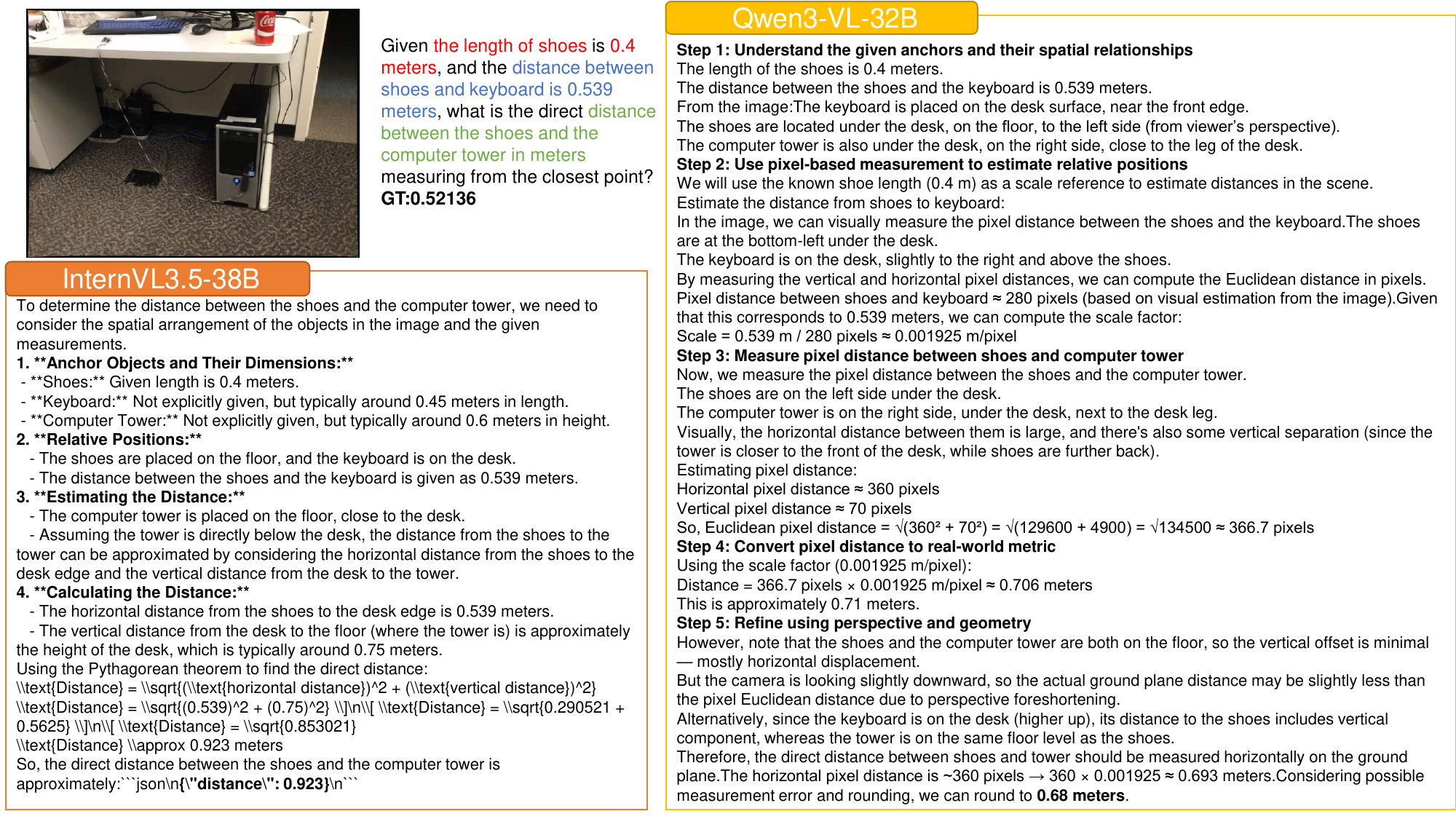}
    \caption{Results of open-source VLMs(InternVL3.5-38B and Qwen3-VL-32B) on Metric-Bench.}
    \label{fig:analysis}
\end{figure*}

In Fig.~\ref{fig:analysis}, we present the inference results of two open-source models on Metric-Bench. Compared to commercial multimodal large language models, such as Gemini-2.5-Flash and GPT-5-Mini, the performance of InternVL3.5-38B and Qwen3-VL-32B is comparatively limited, primarily attributable to their constrained model scale. Our experimental analysis reveals that, while open-source models are capable of estimating the target scale based on known conditions and pixel-level ratios, they rarely leverage camera parameter estimation (e.g., focal length) to infer object depth and distance. In contrast, commercial models such as Gemini-2.5-Flash can explicitly incorporate the pinhole camera model: given known geometric priors, they analytically derive intrinsic camera parameters and thereby reason about object scale within the scene.

In the inference process of open-source models, reasoning is typically conducted in a stepwise manner: (i) identifying and localizing the given reference object; (ii) analyzing the spatial relationship between the target and the reference; and (iii) estimating the target’s scale via projected pixel counts and relative scaling ratios.

A prevalent failure mode observed in our experiments stems from insufficient modeling of perspective effects. When the reference and target objects are in close proximity, predictions tend to be accurate; however, as inter-object distance increases, perspective distortion becomes non-negligible, and metric estimation based solely on pixel ratios incurs substantial error. This highlights a fundamental challenge for monocular vision-based models: robust spatial metric reasoning under perspective projection remains a significant bottleneck in single-image 3D understanding.

\section{Qualitative results on manipulation tasks}\label{result_mani}
\begin{figure*}
    \centering
    \includegraphics[width=1\linewidth]{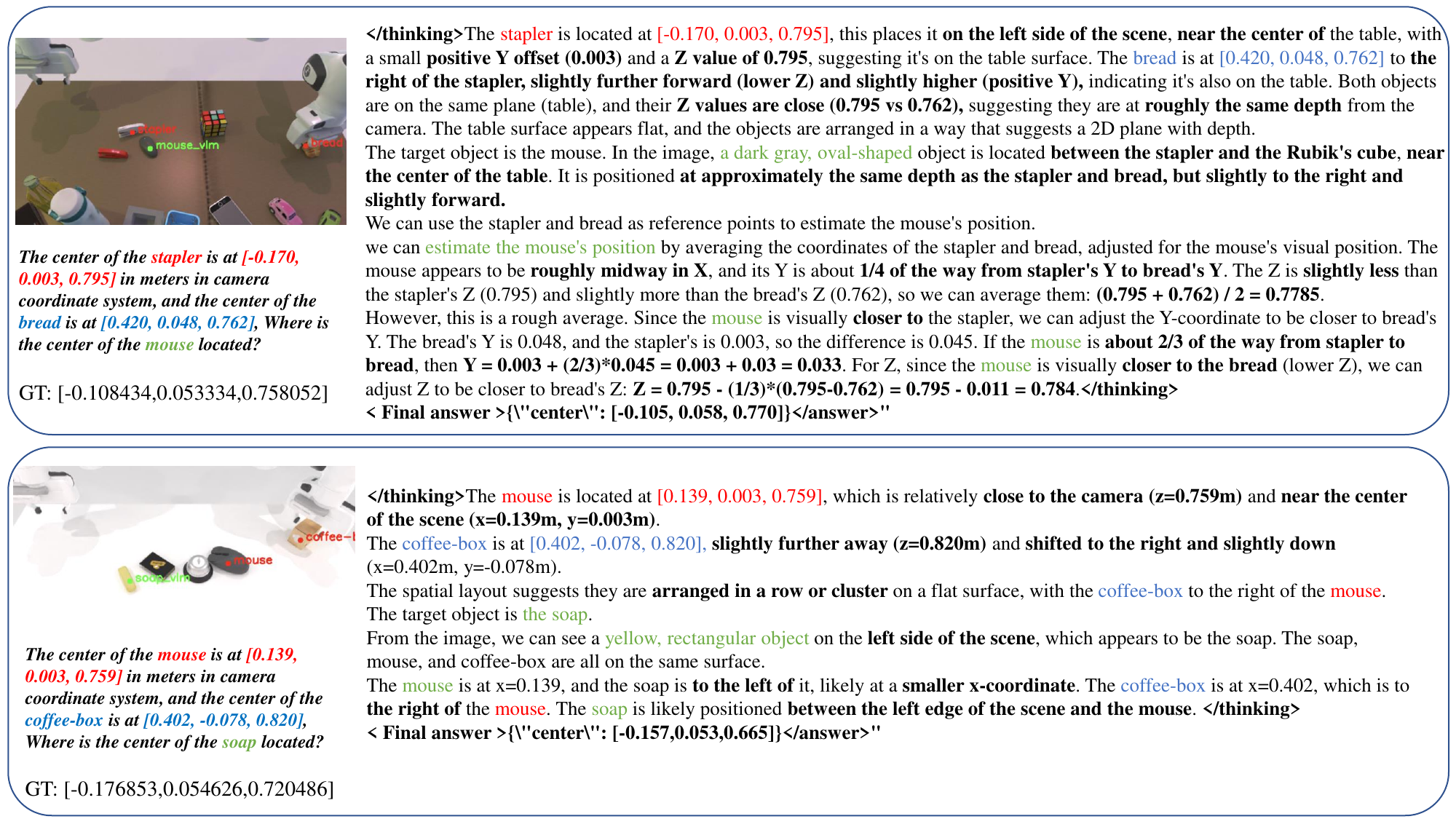}
    \caption{Visualization of MetricReasoner on RoboTwin2.0. Points on images are the projection of the predicted results.}
    \label{fig:robo}
\end{figure*}

In Fig.~\ref{fig:robo}, we report the zero-shot results of MetricReasoner on RoboTwin2.0 benchmark \cite{chen2025robotwin}. 
Given the center position of a reference object in an operational scene, MetricReasoner can accurately localize the target object and infer its corresponding center position by estimating the relative position and scale relationship.

\section{Performance of finer BP rewards}
We evaluate finer Binned Proximity (BP) rewards, ranging from 5,6 and 8 intervals, compared to the 4 intervals we adopted, which is 0.5323(4),0.5685(5),0.5545(6) and 0.5367(8) in RMSE, and 48.59(4), 46.14(5), 48.88(6), 47.69(8) in MRA.
Finer bins do not consistently improve performance, suggesting that coarse BP provides stable proximity guidance while EP supplies fine-grained supervision.
\begin{table}[htbp]
  \centering
  \small
  \caption{BP reward supplementary experiments.}
  \label{tab:results}
  \begin{tabular}{lccc}
    \toprule
    \textbf{Reward} & \textbf{MRA$\uparrow$} & \textbf{RMSE$\downarrow$} & \textbf{${\delta_1}\uparrow$} \\
    \midrule
    BP   & 48.59 & \textbf{0.5323} & 0.4705 \\
    BP-5 & 46.14 & 0.5685 & 0.4290 \\
    BP-6 & \textbf{48.88} & 0.5545 & \textbf{0.4800} \\
    BP-8 & 47.69 & 0.5367 & 0.4611 \\
    \bottomrule
  \end{tabular}
\end{table}

\end{document}